\documentclass{article}
\usepackage{arxiv}

\usepackage{graphicx}
\usepackage{booktabs}
\usepackage{multirow}
\usepackage{amsmath}
\usepackage{url}
\usepackage{enumitem}
\usepackage{hyperref}
\usepackage{caption}
\usepackage{subcaption}
\usepackage{float}
\usepackage{comment}

\title{A Multi-Agent System for Complex Reasoning in Radiology Visual Question Answering}

\author{
  Ziruo Yi \\
  University of North Texas \\
  \texttt{ziruoyi@my.unt.edu}
  \And
  Jinyu Liu \\
  University of North Texas \\
  \texttt{jinyuliu@my.unt.edu}
  \And
  Ting Xiao \\
  University of North Texas \\
  \texttt{ting.xiao@unt.edu}
  \And
  Mark V. Albert \\
   University of North Texas \\
  \texttt{mark.albert@unt.edu}
}

\begin{document}

\maketitle

\begin{abstract}
Radiology visual question answering (RVQA) provides precise answers to questions about chest X-ray images, alleviating radiologists’ workload. While recent methods based on multimodal large language models (MLLMs) and retrieval-augmented generation (RAG) have shown promising progress in RVQA, they still face challenges in factual accuracy, hallucinations, and cross-modal misalignment. We introduce a multi-agent system (MAS) designed to support complex reasoning in RVQA, with specialized agents for context understanding, multimodal reasoning, and answer validation. We evaluate our system on a challenging RVQA set curated via model disagreement filtering, comprising consistently hard cases across multiple MLLMs. Extensive experiments demonstrate the superiority and effectiveness of our system over strong MLLM baselines, with a case study illustrating its reliability and interpretability. This work highlights the potential of multi-agent approaches to support explainable and trustworthy clinical AI applications that require complex reasoning.
\end{abstract}


\section{Introduction}

In modern healthcare, radiology plays a crucial role in diagnosis, outcome prediction, and treatment planning. It relies on diverse data sources such as chest X-ray images, clinical notes, and laboratory tests. Among various multimodal tasks in radiology, radiology visual question answering (RVQA) is particularly valuable as it provides accurate answers to questions about chest X-ray images~\cite{hartsock2024vision} and reduces repetitive tasks, thereby alleviating radiologists’ workload~\cite{yi2025survey}. However, many existing RVQA benchmarks focus on relatively simple question formats and reasoning types. For example, VQA-RAD~\cite{lau2018dataset} and Slake~\cite{liu2021slake} primarily consist of Yes/No or open-ended questions, with limited diversity in task types or reasoning complexity. Although newer datasets such as EHRXQA~\cite{bae2023ehrxqa} offer larger scale and more diverse question templates, the lack of expert-level answer explanations limits their ability to evaluate models' complex reasoning capabilities. While existing benchmarks mainly contain questions requiring only basic reasoning, real-world RVQA scenarios involve subtle visual cues, multi-step inference, and domain knowledge integration, which present significant challenges to the complex reasoning capabilities of current models. This gap underscores the urgent need for RVQA systems that can handle complex, clinically aligned reasoning over multimodal inputs.

With recent advancements in artificial intelligence (AI), computer vision (CV), and natural language processing (NLP), multimodal learning has emerged as a powerful paradigm for integrating and analyzing diverse data sources~\cite{huang2021makes,waqas2024multimodal}. Powered by large language models (LLMs) and large vision models (LVMs) such as GPT-4o~\cite{achiam2023gpt}, LLaMA~3~\cite{meta2024introducing}, and DALL·E~3~\cite{openai2023dalle3}, recent multimodal large language models (MLLMs) have shown promising results across a range of tasks including image captioning~\cite{li2023blip} and visual-language dialogue~\cite{huang2023sparkles}. In the healthcare domain, MLLMs such as Med-PaLM 2~\cite{singhal2025toward} and LLaVA-Med~\cite{li2023llava} have made notable progress in pharmaceutical research~\cite{grisoni2023chemical} and clinical support~\cite{ali2023using}. In RVQA, MLLMs integrate visual and textual information to provide precise answers that support clinical decision-making~\cite{thawkar2023xraygpt, bazi2023vision}. However, existing MLLM approaches for RVQA often treat the model as a black box, combining visual understanding, language grounding, and answer synthesis into a single step. This unified design makes it difficult to interpret intermediate steps, increases the risk of hallucinations under ambiguous cases, and limits the model’s ability to handle complex diagnostic scenarios.

Retrieval-augmented generation (RAG)~\cite{gao2023retrieval,qu2024alleviating,qu2024look} has emerged as a promising strategy to improve the factual accuracy of medical MLLMs. By incorporating external knowledge, RAG enhances contextual understanding and enables more grounded responses. It has been applied to various medical tasks, including report generation~\cite{kumar2024improving, tao2024memory} and visual question answering (VQA)~\cite{yuan2023ramm}. However, applying RAG to RVQA presents new challenges. Retrieving too many contexts can introduce noise and redundancy, while insufficient retrieval may miss key information, ultimately reducing overall answer quality. Given the complexity of RVQA, which demands reliable retrieval, medical reasoning, and cross-modal integration, a single model often struggles to fulfill all requirements. Recently, LLM-based agentic systems have attracted increasing interest due to their ability to perform complex, multi-step tasks through structured collaboration and dynamic interaction~\cite{patil2024gorilla, wang2024survey}. Building on this foundation, multi-agent systems (MASs) have shown potential in areas such as software engineering~\cite{qian2023chatdev, wang2024openhands} and drug discovery~\cite{gottweis2025towards, swanson2024virtual}. However, their application in radiology, especially in RVQA, remains underexplored.

To overcome the limitations of existing MLLM and RAG approaches and to explore the potential of multi-agent systems in radiology, we propose a multi-agent system (MAS) composed of three specialized agents for RVQA: a context understanding agent (CUA), a multimodal reasoning agent (MRA), and an answer validation agent (AVA). This modular design enables structured, stepwise collaboration among agents, enhancing the explainability and precision of the reasoning process while reducing hallucinations. To support our MAS and comprehensively evaluate its complex reasoning capabilities, we curate three subsets from ReXVQA~\cite{pal2025rexvqa}, a large-scale benchmark of multiple-choice questions covering diverse radiological tasks and categories. These subsets enable model disagreement analysis, serve as a retrieval pool for RAG, and facilitate evaluation on challenging cases. The main contributions of our work are as follows:
(1) We present a modular MAS for RVQA that leverages role-specific agents to collaboratively perform complex reasoning tasks, including answer prediction and explanation generation.
(2) We construct three subsets based on ReXVQA to support our MAS and systematically evaluate its performance in challenging scenarios.
(3) We conduct extensive experiments and analyses showing that our MAS consistently outperforms strong MLLM baselines in accuracy, interpretability, and robustness on ambiguous cases.

\section{Related Work}

\textbf{MLLMs for RVQA.} MLLMs have made significant progress in RVQA~\cite{ha2024fusion, li2023masked, gu2024lapa, ossowski2023multimodal, park2024m4cxr, kim2024generalizing}, with models such as ELIXR~\cite{xu2023elixr}, LLaVa-Med~\cite{li2023llava}, and PeFoMed~\cite{liu2024pefomed} combining visual encoders (e.g., ViT~\cite{dosovitskiy2020image}, EVA~\cite{fang2023eva}) with LLMs (e.g., Vicuna~\cite{chiang2023vicuna}, LLaMA2-chat~\cite{touvron2023llama}) to align visual features with textual representations and achieve strong performance on benchmark datasets. Despite their success, these models still suffer from key limitations including reasoning inconsistency~\cite{su2024conflictbank,xia2024cares}, hallucination~\cite{liu2024survey}, and catastrophic forgetting~\cite{khan2023importance,zhai2023investigating}. 
These issues are especially critical in RVQA, where high accuracy and reliability are essential~\cite{wang2025medagent}. In contrast to these black-box approaches, our MAS adopts a modular decomposition with explicit retrieval and validation stages, enhancing reasoning control, interpretability, and robustness.

\textbf{Retrieval-Augmented Generation.} RAG has been increasingly adopted to improve factual grounding in multimodal tasks by retrieving relevant external knowledge~\cite{abootorabi2025ask, mortaheb2025re}. It has been applied to RVQA to reduce hallucinations and enhance factual accuracy~\cite{xia2024mmed}. While RAG offers clear benefits, it still struggles to manage the quantity and quality of retrieved contexts and to mitigate over-reliance on them, which may compromise model performance and introduce factual errors~\cite{xia2024rule}. Moreover, many RAG methods retrieve and process textual and visual information separately, limiting their capacity for integrated multimodal reasoning~\cite{han2025mdocagent}. These limitations are particularly critical in RVQA, where accurate reasoning requires fine-grained alignment between retrieved information and visual evidence. To address these challenges, our system incorporates RAG into a multi-agent workflow, where a dedicated agent retrieves semantically relevant references to support downstream reasoning and validation using both textual and visual inputs. This design enables stronger factual grounding and cross-modal alignment, making it better suited to the complex reasoning demands of RVQA.

\textbf{Multi-Agent Systems.} MASs have gained increasing attention in NLP and healthcare AI~\cite{ke2024enhancing,smit2023we,tang2023medagents,wei2024medco,yue2024ct}. Typically, a MAS consists of a collection of agents that interact through orchestration to enable collective intelligence via coordinated task decomposition, performance parallelization, context isolation, specialized model ensembling, and diverse reasoning discussions~\cite{cemri2025multi, he2025llm, mandi2024roco, zhang2023building, du2023improving, park2023generative, guo2024large}. MASs distribute tasks among specialized agents to collaboratively accomplish complex objectives beyond the scope of single models. While MASs have shown promising results in radiology report generation (RRG)~\cite{alam2025towards,zeng2024enhancing}, their application to RVQA remains largely unexplored. To fill this gap, we propose a multimodal MAS for RVQA that decomposes the complex reasoning process into structured subtasks handled by three specialized agents: a CUA, a MRA, and an AVA. This design facilitates transparent, stepwise collaboration and enables more accurate, robust, and clinically aligned reasoning required in complex radiological scenarios.

\section{Method}

\begin{figure}[!htbp]
  \centering
  \includegraphics[width=\linewidth]{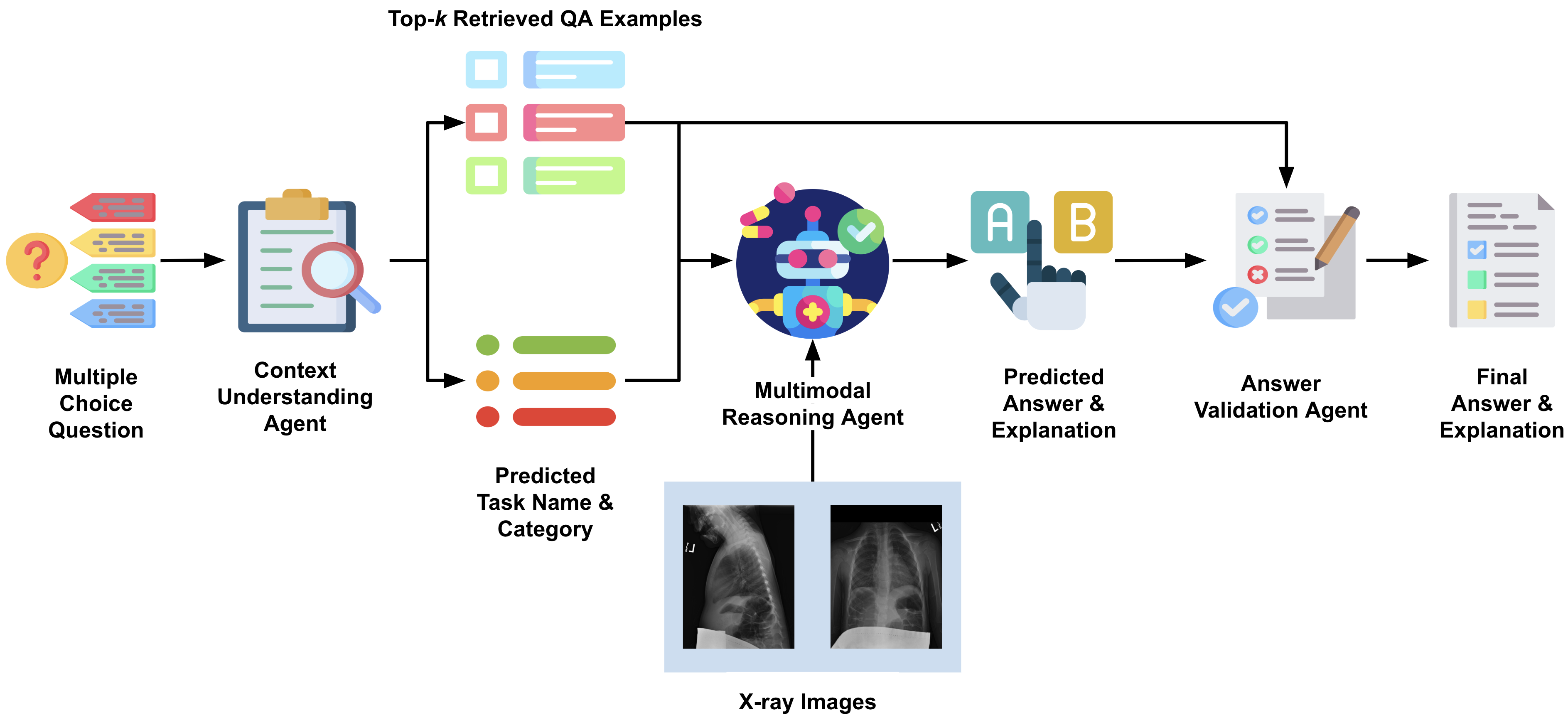} 
  \caption{Overview of the proposed multi-agent system. The pipeline consists of:
(1) a context understanding agent that retrieves top-$k$ relevant QA examples and predicts task name and category;
(2) a multimodal reasoning agent that generates image-grounded answers and explanations; and
(3) an answer validation agent that verifies the predicted answer and revises it when necessary.}
  \label{fig:framework}
\end{figure}

We propose a modular multi-agent system for RVQA that decomposes complex multimodal reasoning into interpretable and cooperative stages. Given a multiple-choice question (MCQ) and one or more corresponding X-ray images, the system sequentially activates three specialized agents: a CUA, a MRA, and an AVA. Each agent serves a distinct functional role and operates independently, using either task-specific prompts (for LLM/MLLM agents) or embedding-based retrieval for selecting top-$k$ relevant examples. As illustrated in Figure~\ref{fig:framework}, the agents communicate through structured intermediate outputs that progressively refine both context understanding and multimodal reasoning into a final answer and explanation. This stepwise architecture enables modular design, interpretability, and flexible integration of different LLMs and MLLMs across agents, aligning with the needs of complex reasoning in radiology MCQs.

\vspace{0.5em}
\vspace{0.25em}

\subsection{Context Understanding Agent}

The CUA supports downstream reasoning by retrieving top-$k$ relevant QA examples and identifying the radiological reasoning task and category of each question. Given an MCQ, we first apply an embedding-based retriever to obtain the top-$n$ most similar QA examples based on sentence-level similarity. These candidates are then reranked by an LLM that scores their relevance to the input MCQ, considering semantic similarity, clinical relevance, and reasoning alignment. We predict the task name and diagnostic category of the input MCQ via weighted voting over the top-$k$ reranked examples. This step provides the predicted task name, category, and top-$k$ QA examples most relevant to the input, which together serve as contextual inputs for the next agent.

\subsection{Multimodal Reasoning Agent}

The MRA generates an answer and explanation by integrating the MCQ, its corresponding X-ray image(s), top-$k$ QA examples, and the predicted task name and category. These elements are passed into a MLLM that leverages both visual evidence and textual knowledge. The MRA then produces a selected answer (A/B/C/D) and a free-text explanation for the choice. This agent produces both the predicted answer and an explanation, together with the top-$k$ QA examples used as inputs for the answer validation step.

\subsection{Answer Validation Agent}

The AVA assesses the reliability of the predicted answer and performs correction if its confidence score falls below a predefined threshold. It receives the MCQ, the top-$k$ QA examples retrieved by the CUA, and the predicted answer and explanation generated by the MRA. The AVA uses an LLM to estimate the confidence score of the predicted answer. If the score exceeds the threshold, the predicted answer and explanation are accepted. Otherwise, the LLM reconsiders the MCQ using the top-$k$ QA examples and generates a revised answer and explanation. This process produces the final output, improving overall answer reliability by correcting low-confidence predictions when necessary.

\section{Experiments}

We conduct a comprehensive evaluation of our multi-agent system on ReXVQA-Hard, a challenging test set from the ReXVQA dataset. Our experiments assess the pipeline's performance in answer accuracy and explanation quality, compare it with strong MLLM baselines, and analyze the contributions of individual agents to complex reasoning in RVQA.

\subsection{Experimental Setup}

\subsubsection{Implementation Details.}

Our system consists of three agents: the CUA, MRA and AVA. The CUA employs Facebook AI Similarity Search (FAISS)~\cite{johnson2019billion} to retrieve the top-$n$ relevant QA examples ($n=10$), which are subsequently reranked by MMed-Llama-3-8B~\cite{qiu2024towards}. We apply a rule-based weighted voting over the top-$k$ examples ($k=5$) to predict the task name and radiological category, assigning greater weight to higher-ranked examples. The MRA is implemented using Med-GEMMA~\cite{sellergren2025medgemma}, an MLLM fine-tuned for medical image and language understanding. MMed-Llama-3-8B is also employed in the AVA for answer validation and correction, using a confidence threshold of 0.7 selected for its consistently strong performance across different values. Since GPT-4o was used to generate the reference answers in the dataset, we exclude GPT-based models from all components of our system to prevent evaluation bias and ensure fairness.

\subsubsection{Datasets.} 

We conduct experiments on ReXVQA, a large-scale benchmark for MCQs in chest radiology. Each example includes one or more chest X-ray images, a clinically meaningful question with four answer options (A/B/C/D), and an expert-written explanation. The dataset covers five radiological reasoning tasks (e.g., presence assessment, differential diagnosis), spanning a diverse range of clinical categories.

\textbf{ReXVQA-Pool.}
We randomly select 600 examples per task, forming a 3,000-example subset that reflects diverse radiological categories. This subset contains a total of 4,795 chest X-ray images and is used to support model disagreement analysis across existing MLLMs.

\textbf{ReXVQA-Hard.}
We identify 1,131 challenging examples from ReXVQA-Pool on which most MLLMs fail to answer correctly, forming a hard subset that serves as the test set for evaluating our multi-agent system.

\textbf{ReXVQA-RAG.}
We construct a separate QA bank of 1,006 diverse examples from the portion of ReXVQA not included in ReXVQA-Pool. This subset is used by the CUA during the retrieval step.

\subsubsection{Baseline Models.}

To evaluate the effectiveness of our multi-agent system, we compare its performance against several state-of-the-art (SOTA) MLLMs on ReXVQA-Hard. These include general-purpose MLLMs (e.g., Janus-Pro-7B~\cite{chen2025janus}, LLaVA 1.5~\cite{liu2023visual}, OpenFlamingo-4B~\cite{awadalla2023openflamingo}, Phi-3.5-Vision-Instruct~\cite{abdin2024phi}, and Qwen2.5-VL~\cite{bai2025qwen2}) as well as a medical-domain MLLM, MedGemma. The selected baselines span a diverse range of architectures and training strategies, providing a representative view of current MLLM capabilities. 
Consistent with our system, we exclude all GPT-based models from the baselines to avoid evaluation bias, since GPT-4o was used to generate the dataset’s reference answers. 
To improve clarity, we refer to the baseline models using the following short names throughout the paper: MedGemma, Janus (Janus-Pro-7B), LLaVA (LLaVA 1.5), OpenFlamingo (OpenFlamingo-4B), Phi35 (Phi-3.5-Vision-Instruct), and Qwen25VL (Qwen2.5-VL). Although Phi-3.5-Vision did not participate in the construction of the ReXVQA-Hard subset, we include it in our evaluation to enable broader comparison across diverse MLLMs.

\subsubsection{Evaluation Metrics.} 

The performance of our MAS is evaluated using accuracy for answer prediction and standard text generation metrics including BLEU~\cite{papineni2002bleu}, ROUGE-L~\cite{lin2004rouge}, METEOR~\cite{banerjee2005meteor}, and BERTScore~\cite{zhang2019bertscore} for explanation quality. These metrics are designed to capture both lexical and semantic similarity between the generated explanations and the ground-truth references, ranging from surface-level overlap (e.g., BLEU, ROUGE-L) to deeper alignment (e.g., METEOR, BERTScore).

\subsubsection{Construction of ReXVQA-Hard}

To construct the challenging test set ReXVQA-Hard, we adopt a difficulty-based filtering strategy inspired by MedAgentsBench~\cite{tang2025medagentsbench}, which highlights the importance of evaluating the complex medical reasoning capabilities in current models. Specifically, we run five MLLMs including MedGemma, Janus, LLaVA, OpenFlamingo, and Qwen25VL on ReXVQA‑Pool, and select examples that are incorrectly answered by at least three of them. This approach ensures that ReXVQA-Hard contains consistently difficult examples across diverse model architectures. As shown in Table~\ref{tab:model_performance}, the five MLLMs exhibit substantial variation in accuracy on ReXVQA‑Pool, ranging from 31.87\% (LLaVA) to 70.77\% (MedGemma). This performance gap suggests that many examples are answered correctly by some models and incorrectly by others, enabling meaningful disagreement-based filtering. By selecting examples that are incorrectly answered by at least three models, we capture questions that are not only difficult for one specific architecture but consistently challenging across models with diverse training approaches. This variation also highlights the limitations of general-purpose MLLMs in handling clinically grounded reasoning tasks, further motivating the need for focused evaluation on hard cases.

\begin{table}[htbp]
\centering
\caption{Accuracy of different MLLMs on ReXVQA-Pool used to construct ReXVQA-Hard.}
\begin{tabular}{lcccccc}
\toprule
\textbf{Metric} & \textbf{MedGemma} & \textbf{Janus} & \textbf{LLaVA} & \textbf{OpenFlamingo} & \textbf{Qwen25VL} \\
\midrule
Accuracy (\%) & 70.77 & 65.30 & 31.87 & 42.40 & 68.27 \\
\bottomrule
\end{tabular}
\label{tab:model_performance}
\end{table}

\subsection{Results}

To evaluate the effectiveness of our MAS, we conduct both quantitative and qualitative analyses on ReXVQA-Hard. This section compares our system with several strong baseline MLLMs in terms of answer accuracy and explanation quality, and presents a case study illustrating how the agents collaborate to resolve diagnostic ambiguity. These results provide a comprehensive assessment of the pipeline’s ability to handle complex reasoning in RVQA.

\subsubsection{Quantitative Analysis}

We evaluate our MAS on ReXVQA-Hard using standard metrics for both answer prediction and explanation generation. As shown in Table~\ref{tab:quant-comparison}, our approach consistently and significantly outperforms recent SOTA MLLMs, including MedGemma, Janus, LLaVA, OpenFlamingo, Phi35, and Qwen25VL. Specifically, our method achieves an accuracy of 63.66\%, outperforming the strongest baseline, MedGemma (44.03\%), by nearly 20 percentage points. This notable improvement highlights our system’s enhanced ability to handle complex and ambiguous examples. In terms of explanation quality, our system obtains the highest scores in BLEU (0.1230) and ROUGE-L (0.3692), suggesting improved lexical alignment with reference explanations. For METEOR, our model achieves a strong score of 0.3449, which is competitive with the best-performing Qwen25VL (0.4125). Our pipeline also achieves a BERTScore of 0.8987, which is comparable to the top score of 0.9008 from Phi35 and reflects strong semantic consistency. These results highlight the effectiveness of integrating contextual understanding, visual reasoning, and answer validation within a MAS. By jointly optimizing for both accuracy and explanation quality, our system generates responses that are not only correct but also clinically meaningful. This makes it particularly suitable for solving difficult MCQs that require complex reasoning in RVQA.

\begin{table}[htbp]
\centering
\caption{Quantitative comparison between our MAS and SOTA MLLMs on ReXVQA-Hard.}
\begin{tabular}{lcccccc}
\toprule
\textbf{Model} & \textbf{Accuracy} & \textbf{BLEU} & \textbf{ROUGE-L} & \textbf{METEOR} & \textbf{BERTScore} \\
\midrule
MedGemma & 44.03\% & 0.0421 & 0.2071 & 0.1903 & 0.8755 \\
Janus & 35.46\% & 0.0192 & 0.2118 & 0.2536 & 0.8691 &  \\
LLaVA & 26.08\% & 0.0048 & 0.0407 & 0.0521 & 0.8171 &  \\
OpenFlamingo & 28.12\% & 0.0612 & 0.2567 & 0.2741 & 0.8798 &  \\
Phi35 & 27.94\% & 0.0802 & 0.3325 & 0.3130 & \textbf{0.9008} &  \\
Qwen25VL & 29.53\% & 0.0750 & 0.2977 & \textbf{0.4125} & 0.8912 &  \\
\textbf{Ours} & \textbf{63.66\%} & \textbf{0.1230} & \textbf{0.3692} & 0.3449 & 0.8987 \\
\bottomrule
\end{tabular}
\label{tab:quant-comparison}
\end{table}
\vspace{0.75em}

\subsubsection{Qualitative Analysis}

\begin{figure}[!htbp]
  \centering
  \includegraphics[width=\linewidth]{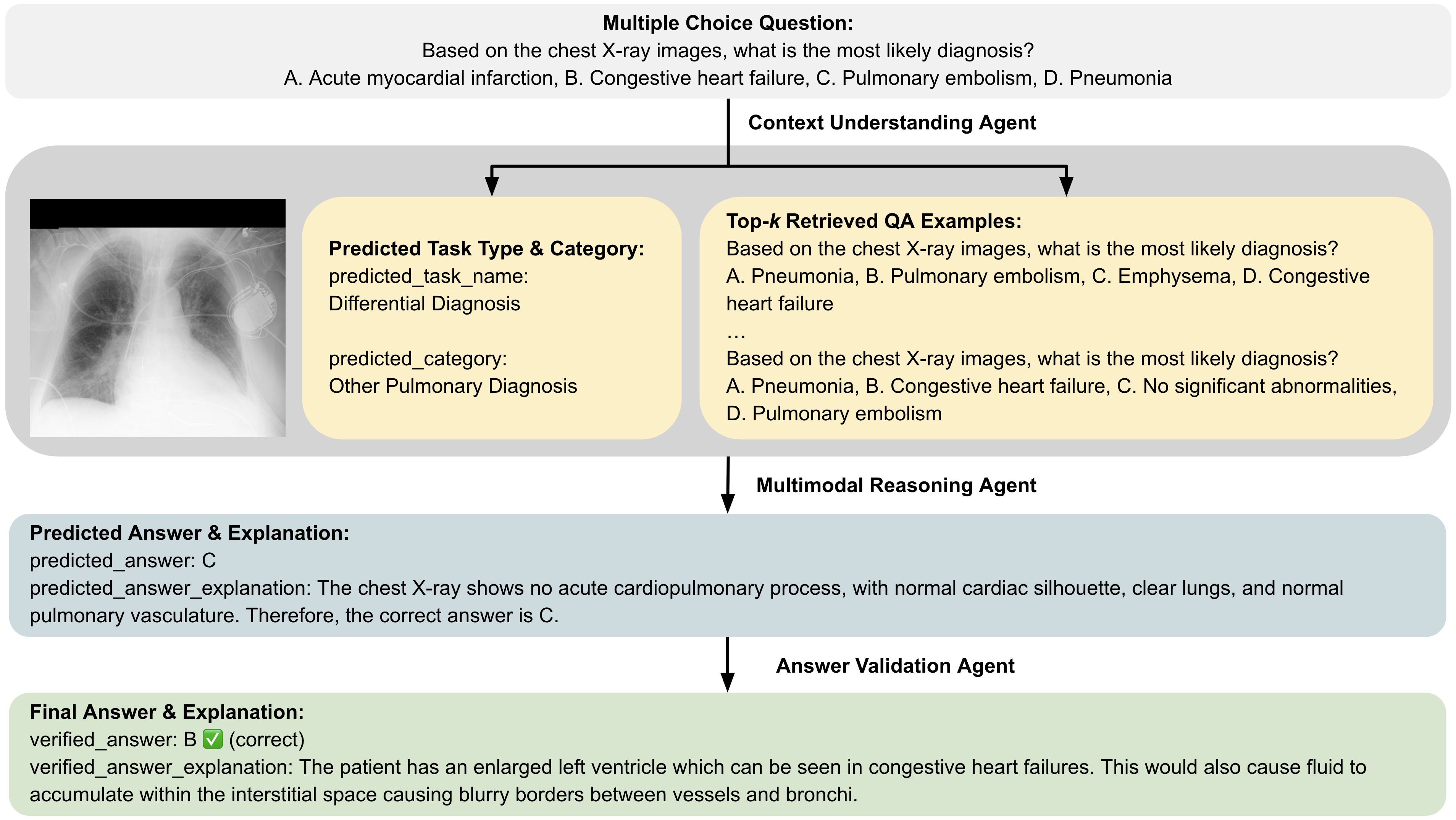}
  \caption{
 A case study showing how the multi-agent system integrates contextual information, visual evidence, and answer validation to support accurate diagnostic reasoning.
  }
  \label{fig:case_study}
\end{figure}

Figure~\ref{fig:case_study} illustrates how our MAS addresses a challenging MCQ. It coordinates contextual understanding, multimodal reasoning, and answer validation to arrive at the correct prediction. The input MCQ asks for the most likely condition based on a chest X-ray, with options including congestive heart failure and pulmonary embolism. The CUA correctly identifies the radiological task name and category, and retrieves QA examples with similar questions and closely related options such as congestive heart failure. The MRA integrates visual and contextual cues to generate a plausible explanation, but ultimately selects an incorrect answer (“C. Pulmonary embolism”), possibly due to the subtle appearance of heart failure on the image. The AVA estimates the confidence of this prediction using an LLM, and determines that it falls below the predefined threshold. It then reconsiders the question using the retrieved QA examples and produces a revised answer: “B. Congestive heart failure,” accompanied by a clinically aligned explanation that supports the diagnosis. This case highlights how contextual knowledge, visual information, and answer validation jointly contribute to resolving diagnostic ambiguity and improving reasoning accuracy.

\subsection{Ablation Study}

\textbf{Agent Contributions to Overall Performance}

To analyze the contribution of each agent, we conduct an ablation study by incrementally enabling them in the pipeline. Specifically, we compare the following three configurations:

1. \textbf{MRA only}: The MRA generates answers and explanations using only the input MCQ and its associated X-ray image(s), without access to retrieved examples or any contextual information.

2. \textbf{CUA + MRA}: The CUA provides retrieved QA examples along with the predicted task name and category, which are used by the MRA in addition to the input MCQ and image(s) to generate an answer and explanation. The AVA is disabled in this configuration, so no verification or correction is applied.

3. \textbf{Full system (CUA + MRA + AVA)}: In the full configuration, the AVA estimates the confidence score of the predicted answer. If the score falls below a predefined threshold, it reconsiders the question using the retrieved QA examples and generates a revised answer and explanation.

The performance of each configuration is shown in Table~\ref{tab:ablation_performance}, with accuracy evaluating answer correctness and BLEU, ROUGE-L, METEOR, and BERTScore assessing explanation quality. Using only the MRA results in the lowest performance across all metrics (e.g., accuracy: 44.03\%, BLEU: 0.0421, BERTScore: 0.8755), indicating that image and question inputs alone are insufficient for reliable reasoning or explanation generation. The lack of contextual guidance limits the model's ability to produce clinically meaningful outputs. Introducing the CUA substantially boosts performance. Accuracy increases by over 10 percentage points (from 44.03\% to 54.29\%), and explanation metrics see similar gains (e.g., BLEU increases from 0.0421 to 0.1078, ROUGE-L from 0.2071 to 0.3532). These results demonstrate that predicted task name, category, and retrieved QA examples are highly effective in guiding both answer selection and explanation generation. Enabling the full system with the AVA further improves performance, achieving 63.66\% accuracy and the highest explanation scores across all metrics (e.g., BLEU: 0.1230, BERTScore: 0.8987). These improvements highlight the value of answer validation in identifying and correcting low-confidence predictions, ultimately improving output accuracy and trustworthiness. Overall, each component plays a critical role in system performance. The full pipeline benefits from contextual retrieval, multimodal reasoning, and answer validation, highlighting the importance of a modular and cooperative architecture for complex reasoning in RVQA.

\begin{table}[htbp]
\centering
\caption{Performance comparison of different agent configurations on ReXVQA-Hard.}
\begin{tabular}{lcccccc}
\toprule
\textbf{Agents} & \textbf{Accuracy} & \textbf{BLEU} & \textbf{ROUGE-L} & \textbf{METEOR} & \textbf{BERTScore} \\
\midrule
MRA only & 44.03\% & 0.0421 & 0.2071 & 0.1903 & 0.8755 \\
CUA + MRA & 54.29\% & 0.1078 & 0.3532 & 0.3225 & 0.8950 \\
\textbf{Full System} & \textbf{63.66\%} & \textbf{0.1230} & \textbf{0.3692} & \textbf{0.3449} & \textbf{0.8987} \\
\bottomrule
\end{tabular}
\label{tab:ablation_performance}
\end{table}

\subsection{Discussion}

Our experiments demonstrate that the proposed MAS significantly improves performance on ReXVQA-Hard, a test set selected based on model disagreement to include consistently difficult examples. By combining contextual retrieval, multimodal reasoning, and answer validation, our system consistently outperforms strong baselines across both answer accuracy and explanation quality. These results highlight the importance of modular and interpretable reasoning pipelines for tackling radiological questions that require complex reasoning.

The CUA enriches contextual information by retrieving semantically relevant QA examples and predicting the radiological task type and category, which provides critical guidance to downstream components. The MRA then fuses visual and textual modalities to generate clinically plausible answers and explanations, effectively addressing the complexity of radiological reasoning. Finally, the AVA estimates the confidence of each prediction and revises low-confidence answers when necessary, improving reliability on clinically ambiguous cases.

The integration of the three specialized agents results in a modular system that balances performance, explainability, and clinical alignment. The ablation study confirms that each agent plays a key role in the system’s overall performance, while the case study illustrates how their coordination improves diagnostic reasoning in challenging cases. One limitation is that the AVA relies on a fixed confidence threshold to trigger answer revision, which may not generalize well across different medical domains. To address this, future work will explore dynamic thresholding strategies or learned validation mechanisms to enhance flexibility and robustness, especially for difficult or rare cases.

\section{Conclusion}

We present a MAS for RVQA, which decomposes the task into three specialized agents for context understanding, multimodal reasoning, and answer validation. This modular design fully leverages multimodal information and enables interpretable and step-wise reasoning. Our system significantly outperforms strong MLLM baselines, achieving notable improvements in both answer accuracy and explanation quality. The ablation study demonstrates the importance of each agent, while qualitative analysis illustrates how their collaboration resolves diagnostic ambiguity. This pipeline provides a flexible and generalizable approach for other multimodal medical tasks that require complex reasoning and clinical precision.

\bibliographystyle{unsrt}
\bibliography{template}

\end{document}